\PassOptionsToPackage{colorlinks=true,linkcolor=black,citecolor=black,urlcolor=black}{hyperref}
\documentclass[conference]{IEEEtran}
\IEEEoverridecommandlockouts

\usepackage{soul}
\usepackage{xcolor}
\usepackage{color, soul}
\usepackage{amsmath}
\usepackage{amsmath,amssymb,amsfonts}
\usepackage{algorithmic}
\usepackage{graphicx}
\usepackage{textcomp}
\usepackage{orcidlink} 
\usepackage[none]{hyphenat}
\usepackage{booktabs} 
\usepackage{multirow} 
\usepackage[colorlinks=true,linkcolor=black,citecolor=black,urlcolor=black]{hyperref} 

\def\BibTeX{{\rm B\kern-.05em{\sc i\kern-.025em b}\kern-.08em
    T\kern-.1667em\lower.7ex\hbox{E}\kern-.125emX}}

\begin{document}

\IEEEpubid{\makebox[\columnwidth]{\parbox{\columnwidth}{
\vspace{30pt}\fontsize{7}{8}\selectfont
\copyright~2026 IEEE. Personal use of this material is permitted. Permission from IEEE must be obtained for all other uses, in any current or future media, including reprinting/republishing this material for advertising or promotional purposes, creating new collective works, for resale or redistribution to servers or lists, or reuse of any copyrighted component of this work in other works.}
\hfill} \hspace{\columnsep}\makebox[\columnwidth]{ }}

\title{XMedFusion: A Knowledge-Guided Multimodal Perception and Reasoning Framework for Autonomous Medical Systems}

\author{
\IEEEauthorblockN{
Hamza Riaz\,\orcidlink{0009-0005-5370-8576}\IEEEauthorrefmark{1},
Arham Haroon\,\orcidlink{0009-0008-3046-6633}\IEEEauthorrefmark{1},
Maha Baig\,\orcidlink{0009-0004-4249-5088}\IEEEauthorrefmark{1},
Muhammad Dawood Rizwan\,\orcidlink{0009-0005-8971-3035}\IEEEauthorrefmark{1},\\
Muhammad Naseer Bajwa\,\orcidlink{0000-0002-4821-1056}\IEEEauthorrefmark{1},
Muhammad Moazam Fraz\,\orcidlink{0000-0003-0495-463X}\IEEEauthorrefmark{1},\IEEEauthorrefmark{2}
}

\IEEEauthorblockA{\IEEEauthorrefmark{1}
National University of Sciences and Technology (NUST), Islamabad, Pakistan
}

\IEEEauthorblockA{\IEEEauthorrefmark{2}University of Staffordshire, Stoke-on-Trent, ST4 2DE, United Kingdom}

\thanks{$^{*}$ Equal contribution: Hamza Riaz, Arham Haroon, Maha Baig}

}

\maketitle

\begin{abstract}
Autonomous medical and robotic systems increasingly rely on intelligent perception and reasoning capabilities to interpret visual data and support clinical decision making. Radiology report generation represents a critical component of such automated diagnostic workflows, yet existing end-to-end multimodal models often suffer from weak visual grounding, resulting in unreliable interpretations and omission of subtle clinical findings. This paper presents XMedFusion, a modular AI framework designed as an intelligent perception and reasoning module for autonomous medical systems. The proposed framework decomposes visual information into coordinated functional components that emulate expert-driven analysis, including a visual perception agent that extracts image-grounded evidence, a knowledge graph construction agent that structures clinically relevant findings, and a retrieval-guided drafting process that ensures a consistent reporting structure. A synthesis agent iteratively integrates visual and structured evidence through reasoning-driven verification to produce reliable and interpretable diagnostic outputs. Experimental evaluation on a public chest radiograph dataset demonstrates significant improvements over baseline vision-language models, achieving gains from 0.0493 to 0.3359 in BLEU-1, 0.0863 to 0.2440 in ROUGE-L, and 0.0829 to 0.1708 in METEOR, along with substantial improvements in semantic evaluation metrics such as Consistency (2.38 to 7.80) and Accuracy (2.34 to 6.93). The results highlight the effectiveness of structured multi-agent perception and reasoning for enhancing robustness, transparency, and automation in intelligent medical imaging systems, enabling  integration into autonomous healthcare and robotic diagnostic workflows.
\end{abstract}


\section{Introduction}

As autonomous medical and robotic systems become more capable, the need for reliable perception and reasoning over complex visual data is becoming increasingly important for clinical decision support. Radiology workflows represent a challenging instance of such multimodal perception tasks, where medical images must be analyzed together with clinical context to produce accurate diagnostic interpretations. In this setting, Radiology Report Generation (RRG) aims to automatically produce clinically meaningful reports that mirror expert radiologist documentation and enable reliable downstream decision making~\cite{yildirim2024multimodal, yi2025survey}. Most RRG approaches focus on chest X-ray imaging due to its widespread use and the importance of structured narrative reporting in clinical communication. However, increasing imaging volumes and persistent shortages of radiologists continue to impose significant workload burdens and reporting delays~\cite{mcdonald2015effects}. These limitations highlight the need for automated perception and reasoning modules that can enhance diagnostic efficiency and support autonomous healthcare and robotic diagnostic workflows~\cite{petinaux2011accuracy}.

\IEEEpubidadjcol 

Recent advances in multimodal large language models (MLLMs) have enabled vision-language systems to process medical images through visual encoders and generate textual interpretations, accelerating end-to-end approaches to RRG. These models jointly reason over visual and textual information and can refine outputs during inference. However, reliable clinical interpretation requires more than fluent text generation. The system must capture subtle visual evidence, map findings to clinically grounded concepts, and produce precise diagnostic descriptions. In practice, MLLMs often suffer from weak visual grounding, where missed fine-grained abnormalities or uncertain visual representations cause the model to rely on learned priors, resulting in hallucinated findings and reduced diagnostic reliability. Such limitations motivate structured reasoning designs, as intermediate reasoning steps have been shown to improve consistency and interpretability in language models.

\begin{figure*}[t]
    \centering
    \includegraphics[width=\textwidth]{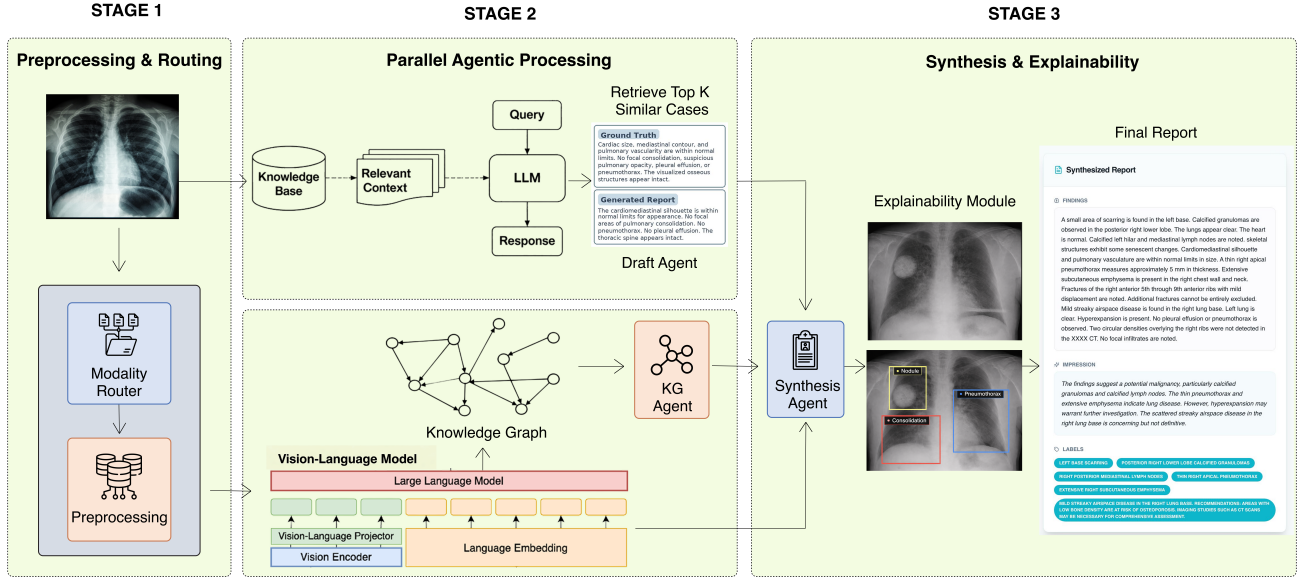} 
\caption{Architecture of the XMedFusion framework. The system consists of four coordinated agents (Vision, Knowledge Graph, Retrieval and Draft, and Synthesis) that perform multimodal perception, structured reasoning, and evidence-grounded decision making for autonomous medical imaging and diagnostic workflows.}  \label{fig:pipeline}
\end{figure*}

To address these challenges, automated diagnostic systems must move beyond single-pass image-to-text mappings toward structured perception pipelines that externalize evidence, perform interpretable reasoning, and enable verification of intermediate outputs. In this work, we propose \textbf{XMedFusion}, a knowledge-guided multimodal perception and reasoning framework that models RRG as a coordinated multi-agent process operating over interpretable intermediate representations. The proposed framework decomposes visual interpretation into complementary components, including a visual perception agent that extracts image-grounded evidence, a knowledge graph construction agent that structures clinically relevant findings, and a retrieval-guided drafting mechanism that provides consistent diagnostic reporting structure. A synthesis agent integrates these intermediate representations through iterative reasoning to generate reliable and interpretable diagnostic outputs. By explicitly grounding perception and reasoning in structured evidence, XMedFusion improves cross-modal consistency, reduces hallucinations, and enhances transparency, enabling its integration as an intelligent perception agent for autonomous medical imaging and robotic diagnostic systems. The overall framework is illustrated in Fig.~\ref{fig:pipeline}.

The main contributions of this work are as follows:
\begin{itemize}
\item We propose XMedFusion, a knowledge-guided multi-agent framework for multimodal medical image interpretation that decomposes report generation into structured perception, knowledge representation, and evidence-grounded synthesis components.

\item We design a neuro-symbolic reasoning pipeline that externalizes visual evidence through knowledge graph modeling, improving cross-modal alignment and reducing hallucinations in vision-language models.

\item We introduce an iterative synthesis mechanism that integrates visual descriptions, structured knowledge, and retrieved contextual priors to generate clinically consistent and interpretable reports.

\item We demonstrate the effectiveness of the proposed framework on a public chest radiograph benchmark, achieving significant improvements in both lexical and semantic evaluation metrics over existing approaches.
\end{itemize}

\section{Related Work}

\subsection{Multimodal LLMs for Medical Image Understanding}

Recent advances in multimodal large language models (MLLMs) have enabled intelligent perception systems that jointly process visual and textual information for automated medical image interpretation. Existing RRG approaches typically couple medical vision encoders with pretrained language models through alignment modules that map visual features into language representations. Representative methods include \emph{R2GenGPT}~\cite{wang2023r2gengpt}, which employs a frozen language model with lightweight visual alignment, and \emph{XrayGPT}, which integrates a medical vision-language encoder with a conversational decoder for radiograph analysis. Domain-adapted systems such as \emph{MAIRA-1} further demonstrate that specialized visual encoders improve diagnostic interpretation quality~\cite{srivastav2024maira}. Despite these advances, end-to-end MLLM-based systems often suffer from weak visual grounding, where missed or poorly localized findings lead to hallucinated interpretations and reduced reliability~\cite{xia2024cares}. These limitations limit their deployment in autonomous diagnostic and robotic perception workflows that require robust and interpretable visual understanding.

\subsection{Retrieval-Augmented Diagnostic Report Generation}

Retrieval-augmented generation methods improve interpretability and consistency by incorporating prior knowledge during visual interpretation~\cite{abdullah2025multi}. In medical imaging, retrieval-based approaches identify similar cases using multimodal embeddings and condition prediction on retrieved contextual information. Structured retrieval strategies further leverage clinical knowledge schemas such as RadGraph to preserve factual consistency in generated outputs~\cite{jain2021radgraph}. However, retrieval-based pipelines remain sensitive to retrieval quality and may introduce irrelevant or conflicting information, leading to unreliable predictions~\cite{xia2024rule}. Moreover, retrieved context does not inherently guarantee consistency with current visual observations, highlighting the need for structured reasoning mechanisms that explicitly reconcile visual evidence with contextual knowledge.

\subsection{Agent-Based and Multi-Agent Perception Systems}

Agent-based and multi-agent structures have surfaced as a viable approach for autonomous systems by decomposing complex perception and reasoning tasks into modular components with intermediate verification. In medical AI, frameworks such as \emph{MDAgents} coordinate multiple language models to emulate expert consultation processes~\cite{kim2024mdagents}, while \emph{MMedAgent} enables adaptive selection of specialized tools for multimodal medical analysis~\cite{li2024mmedagent}. Systems such as \emph{MedRAX} further demonstrate the integration of multimodal reasoning and tool-based workflows for automated chest radiograph analysis. While these techniques improve modular reasoning, ensuring consistency between visual perception and generated outputs remains challenging. In particular, existing methods do not explicitly combine image-grounded evidence, structured clinical representation, and contextual guidance within a unified framework, which limits reliability in autonomous diagnostic and robotic imaging systems.

Recent work on reasoning with medical LLMs has further highlighted the importance of graph-based reasoning, grounded retrieval, and hybrid neuro-symbolic pipelines for deployment in high-stakes clinical settings~\cite{mansoor2026reasoning}. Despite recent progress, a key gap remains in RRG. End-to-end multimodal models generate fluent reports but do not explicitly externalize the visual evidence behind their predictions. Retrieval-augmented methods improve factual consistency, yet remain sensitive to retrieval quality and may introduce conflicting context. Medical agent systems support modular reasoning and tool use, but generally do not use structured clinical representations as control signals during report synthesis.

In contrast, XMedFusion combines dense image-grounded evidence extraction, neuro-symbolic knowledge graph construction, retrieval used only as contextual scaffolding, and evidence-prioritized iterative synthesis. This addresses a distinct gap at the intersection of visual grounding, structured clinical representation, and verification-driven report generation.

\section{Methodology}
\label{sec:methodology}

The proposed framework is designed as a modular perception and reasoning architecture for autonomous medical imaging and robotic diagnostic systems. Rather than relying on single-pass image-to-text generation, the system decomposes visual interpretation into structured stages that externalize evidence, construct intermediate representations, and perform verification prior to final decision synthesis. This staged design improves robustness, interpretability, and reliability, enabling integration into autonomous diagnostic workflows.

XMedFusion consists of four coordinated agents operating in parallel: the Vision Agent, the Knowledge Graph (KG) Agent, the Retrieval and Draft Agent, and the Synthesis Agent. Each agent performs a specialized function within the perception--reasoning pipeline. The Vision Agent extracts dense image-grounded descriptions, the KG Agent constructs structured knowledge representations, the Retrieval and Draft Agent provides contextual guidance and initial structure, and the Synthesis Agent integrates these components through iterative reasoning. Although the framework supports multi-modality through a routing mechanism, this work focuses on chest X-ray radiographs due to dataset availability issues.

\subsection{Problem Formulation}
Given an input medical image $x$, the goal is to generate a clinically accurate diagnostic report $y$. Instead of directly modeling the conditional distribution $P(y \mid x)$, our proposed framework decomposes the generation process into structured intermediate representations.

The Vision Agent produces an image-grounded description:
\begin{equation}
d_{\text{vis}} = f_{\text{vis}}(x)
\end{equation}

The Knowledge Graph Agent constructs a structured representation:
\begin{equation}
G = (V, E)
\end{equation}
where $V$ represents entities (anatomy and observations) and $E$ represents relations between them.

The Retrieval and Draft Agent retrieves similar cases:
\begin{equation}
s_i = \cos(v, r_i), \quad D_k = \text{TopK}(s_i, k)
\end{equation}
where $v$ is the image embedding and $r_i$ are retrieved case embeddings.

Finally, the Synthesis Agent generates the report:
\begin{equation}
\hat{y} = \arg\max_y P_\theta(y \mid d_{\text{vis}}, G, D_k)
\end{equation}

This approach keeps perception, reasoning, guidance, and report generation distinct. We describe the role of each agent in the following subsections.

\subsection{Vision Agent}

The Vision Agent functions as the primary visual perception agent. It analyzes the input radiograph and generates a dense, image-grounded description covering clinically relevant anatomical regions such as lung fields, pleura, mediastinum, and cardiac silhouette. The objective of this agent is to externalize visual evidence into an explicit intermediate representation that can be verified and reused by downstream reasoning components. The module is constrained to describe only directly observable findings, reducing speculative interpretation and improving perceptual reliability.

\subsection{Knowledge Graph Agent}

To ensure structured reasoning and factual consistency, the KG Agent converts visual evidence into an explicit knowledge graph representation using a hybrid neuro-symbolic pipeline. A spatially aware detection mechanism based on BioMedCLIP~\cite{zhang2023biomedclip} processes the image at both global and regional levels. A multi-label classifier establishes global pathology priors, while region-specific crops are analyzed using zero-shot prompting to localize findings.

\begin{figure}[h]
    \centering
    \includegraphics[width=0.95\linewidth]{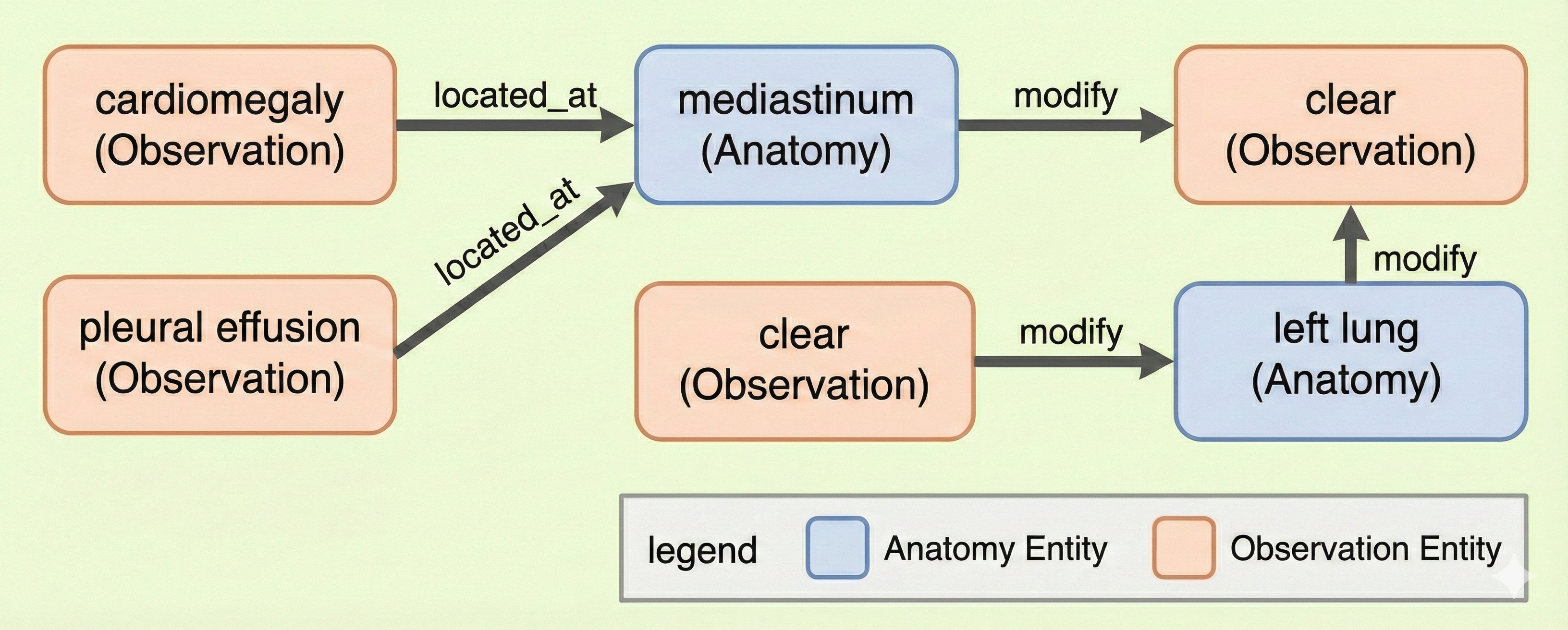}
    \caption{Example knowledge graph generated by the Knowledge Graph Agent. Visual findings are encoded as structured entities and relations linked to anatomical regions, providing an interpretable intermediate representation for structured reasoning and evidence-grounded decision synthesis in autonomous medical imaging systems.}
    \label{fig:kg_diagram}
\end{figure}

A logical gating mechanism enforces consistency between global and local predictions. When the global classifier predicts the absence of a pathology with high confidence, low-confidence regional detections are suppressed, reducing false positives. Validated findings are mapped to a predefined ontology of thoracic pathologies and instantiated within a RadGraph-compliant schema~\cite{jain2021radgraph}. The resulting graph consists of \textit{Anatomy} and \textit{Observation} nodes connected through relational edges. Negative findings are explicitly encoded, enabling the representation to function as a strict control signal for downstream synthesis and preventing unsupported diagnostic statements. An example of the structured knowledge representation produced by the KG Agent is shown in Fig.~\ref{fig:kg_diagram}.

\subsection{Retrieval and Draft Agent}

The Retrieval and Draft Agent enhances contextual consistency by incorporating prior cases during diagnostic interpretation. The input image is projected into a shared embedding space and compared against a database of report representations using cosine similarity. The top-$k$ nearest cases are selected to provide structural and stylistic priors.

Based on the retrieved cases, an initial diagnostic scaffold is generated to establish a coherent reporting structure. This scaffold identifies common patterns from similar cases as a starting point, though every detail still requires verification. Retrieved information is treated strictly as contextual guidance rather than evidence, ensuring that final decisions remain grounded only in current visual observations.

\subsection{Synthesis Agent}

The Synthesis Agent generates the final diagnostic report by consolidating grounded visual descriptions, structured knowledge, and contextual drafts. Drawing on iterative reasoning and self-refinement paradigms~\cite{yao2022react}, the agent continuously evaluates how well the report aligns with the visual evidence. This process allows it to identify and resolve any inconsistencies, ensuring the final output remains strictly evidence-based. Visual perception signals and structured knowledge representations are prioritized over retrieval-based priors during conflict resolution. This iterative verification mechanism ensures cross-modal consistency, reduces hallucinations, and produces reliable diagnostic outputs suitable for autonomous system deployment.

XMedFusion introduces novelty by assigning distinct evidential responsibilities to each of its four modules, ensuring a structured transition from perception to reasoning. The Vision Agent externalizes observable findings as explicit evidence, while the Knowledge Graph Agent extracts structured clinical representations from the same scan. The Retrieval and Draft Agent contributes contextual priors for guidance, enabling the Synthesis Agent to consolidate all information through evidence-prioritized reasoning. This explicit separation of roles distinguishes XMedFusion from conventional retrieval-centric pipelines and general multi-agent systems.

\section{Experiments}

The experimental evaluation assesses the effectiveness of the proposed multimodal perception and reasoning framework for autonomous medical imaging systems. The evaluation focuses on system reliability, semantic consistency, and robustness compared to conventional single-pass vision-language models.

\subsection{Experimental Setup}

\subsubsection{Implementation Details}

The proposed system integrates four coordinated modules: Vision, Knowledge Graph, Retrieval and Draft, and Synthesis. The retrieval component follows the RULE methodology~\cite{xia2024rule}, employing a CLIP-based encoder fine-tuned on the IU X-ray dataset using contrastive learning to align visual and textual representations. The top-$k$ most similar cases ($k=3$) are selected using cosine similarity.

For visual perception, the Vision Agent utilizes BioMedCLIP~\cite{zhang2023biomedclip} with a projection layer to generate dense image-grounded descriptions. The Knowledge Graph Agent shares the same encoder to construct structured entity--relation representations validated against a predefined schema. All generative modules are powered by MedGemma 1.5:4B~\cite{sellergren2025medgemma}, enabling deployment in resource-constrained environments. During synthesis, image-grounded evidence is prioritized over retrieval-based priors to improve consistency and reduce hallucinations.

\subsubsection{Dataset}

Evaluation is conducted on the IU X-ray dataset \cite{demner2016preparing}, a publicly available benchmark for RRG. The dataset consists of chest radiograph studies paired with corresponding expert-written reports, where each report includes \textit{Findings} and \textit{Impression} sections describing observed abnormalities and clinical conclusions. Each study contains multiple radiographic views (e.g., frontal and lateral). Following common practice, we use only the frontal view image for each study to ensure consistent single-image report generation.

Following prior work \cite{liu2024factual}, we use a preprocessed subset of the dataset consisting of 2,068 image-report pairs for training and 590 pairs for testing. Images are resized and normalized before being processed by the visual encoder. Reports are used in their original form, allowing evaluation of both factual correctness and natural language generation quality. The training split is used to construct the retrieval database by encoding image-report pairs into a shared embedding space. During inference, only the test split is used for evaluation, ensuring no overlap between retrieval candidates and test samples.

\subsubsection{Evaluation Metrics}

We evaluate the generated reports using both lexical similarity and semantic reliability metrics. BLEU measures modified n-gram precision with a brevity penalty to account for overly short outputs.

ROUGE-L measures the longest common subsequence (LCS) between generated and reference reports, capturing sentence-level structural similarity.

METEOR evaluates alignment based on a harmonic mean of precision and recall, with additional consideration for synonym and stem matching \cite{banerjee2005meteor}.

In addition, we also employ an LLM-as-a-Judge framework \cite{zheng2023judging} to assess semantic quality across Coverage, Consistency, Accuracy, Style, and Conciseness. This provides a higher-level evaluation of factual correctness and reasoning fidelity, complementing traditional n-gram based metrics.

\section{Results}

\subsection{Quantitative Performance}

We evaluate the proposed XMedFusion framework on the IU X-ray test set and compare its performance with LLaVA-Med 1.5 \cite{li2023llava}, which serves as a very capable baseline vision-language model. Quantitative results are summarized in Table~\ref{tab:metric_comparison}.

The proposed framework demonstrates substantial improvements across all evaluation metrics. The increase in BLEU-1 and ROUGE-2 indicates stronger phrase-level consistency and improved structural alignment with reference reports, while gains in ROUGE-L and METEOR reflect enhanced semantic coherence and output quality. These results show that structured perception and reasoning improve the reliability of automated medical image interpretation compared to single-pass generation approaches.

\begin{table}[t]
\centering
\caption{Comparison between LLaVA-Med and XMedFusion on the IU X-ray test set. Best results are shown in bold.}
\label{tab:metric_comparison}
\begin{tabular}{lcc}
\toprule
\textbf{Metric} & \textbf{LLaVA-Med} & \textbf{XMedFusion} \\
\midrule
BLEU-1  & 0.0493  & \textbf{0.3359} \\
ROUGE-1 & 0.1150  & \textbf{0.2385} \\
ROUGE-2 & 0.0213  & \textbf{0.1328} \\
ROUGE-L & 0.0863  & \textbf{0.2440} \\
METEOR  & 0.0829  & \textbf{0.1708} \\
\bottomrule
\end{tabular}
\end{table}

\subsection{Semantic Reliability Evaluation}

To assess reasoning fidelity and evidence alignment, we perform semantic evaluation using the LLM-as-a-Judge framework. The results are presented in Table~\ref{tab:llm_judge}.

XMedFusion achieves significantly higher scores across all evaluation dimensions, particularly in Accuracy and Consistency, demonstrating improved reasoning reliability and reduced hallucinations. The improvements confirm that structured knowledge representation and evidence-grounded verification enhance diagnostic robustness.

\subsection{Explainability and System Transparency}

\begin{table}[h]
\centering
\caption{LLM-as-a-Judge evaluation (scale 1--10).}
\label{tab:llm_judge}
\begin{tabular}{lcc}
\toprule
\textbf{Metric} & \textbf{LLaVA-Med} & \textbf{XMedFusion} \\
\midrule
Coverage & 2.54 & \textbf{5.73} \\
Consistency & 2.38 & \textbf{7.80} \\
Accuracy & 2.34 & \textbf{6.93} \\
Style & 3.22 & \textbf{7.00} \\
Conciseness & 2.16 & \textbf{7.33} \\
\bottomrule
\end{tabular}
\end{table}

\begin{figure}[htbp]
\centering
\includegraphics[width=0.75\linewidth]{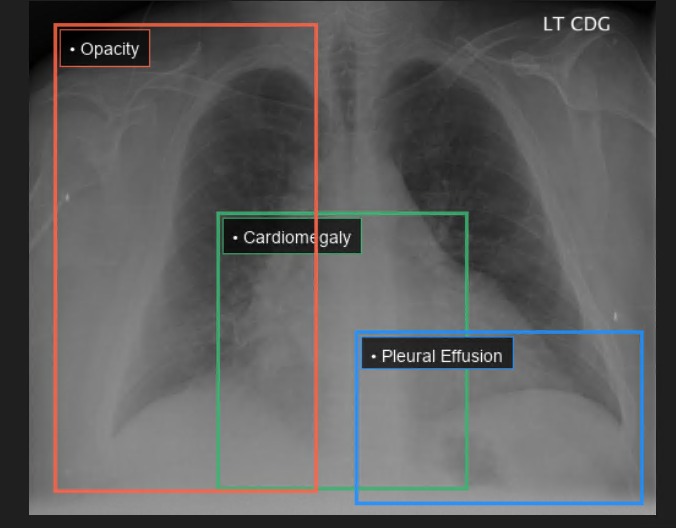}
\caption{Explainability module of XMedFusion showing structured visual grounding and interpretable reasoning for autonomous medical imaging systems.}
\label{fig:explainability}
\end{figure}

The framework employs a visual grounding mechanism that maps extracted knowledge graph entities directly to anatomical regions within the scan. As shown in Fig.~\ref{fig:explainability}, the system generates visual overlays to explicitly demonstrate what the agents identify in the input scan. This transparency allows radiologists to verify the agent's findings by cross-referencing the generated report with the highlighted regions in the scan. By providing a traceable path from visual perception to final synthesis, this neuro-symbolic grounding ensures that autonomous diagnostic workflows remain interpretable and clinically reliable.

\subsection{Discussion}

The results show that decomposing visual interpretation into modular perception and reasoning stages significantly improves performance compared to single-pass vision-language models. The combination of explicit visual grounding, structured knowledge representation, and iterative verification enhances semantic consistency, reduces hallucination, and improves system robustness. The Vision Agent strengthens perceptual grounding by making image evidence explicit, the Knowledge Graph Agent improves factual consistency through structured representation, the Retrieval and Draft Agent helps organize the report, and the Synthesis Agent keeps generation aligned with the strongest available evidence. The large gains in Consistency and Accuracy further suggest that structured knowledge and evidence-driven synthesis are especially important for diagnostic reliability.

\section{Conclusion}

This paper presented XMedFusion, a knowledge-guided multimodal perception and reasoning framework for autonomous medical imaging and robotic diagnostic systems. The framework decomposes visual interpretation into structured perception, knowledge representation, contextual retrieval, and iterative synthesis, enabling reliable and interpretable diagnostic report generation. By grounding visual evidence in structured representations with verification, XMedFusion improves cross-modal consistency, reduces hallucinations, and enhances robustness over single-pass VLMs. Results on the IU X-ray dataset show clear gains across both lexical and semantic metrics, confirming stronger diagnostic reliability and reasoning fidelity. The integrated neuro-symbolic explainability module further provides transparent visual grounding and traceability, supporting trustworthy deployment in autonomous healthcare and perception systems. However, evaluation is currently limited to a single chest radiograph dataset. Future work will extend to additional imaging modalities and real-world autonomous diagnostic workflows.

\section*{Acknowledgment}
This work has received funding from the German Academic Exchange Service (DAAD)
under the Project ID: 8979614, titled \textit{Ba-Ikhtiyar Jawan: Upscaling and Digitization of Vocational Education Curriculum}. AI was used only for editorial assistance. All scientific content is the authors’ original work.

\bibliographystyle{IEEEtran}
\bibliography{reference}

\begin{thebibliography}{10}
\providecommand{\url}[1]{#1}
\csname url@samestyle\endcsname
\providecommand{\newblock}{\relax}
\providecommand{\bibinfo}[2]{#2}
\providecommand{\BIBentrySTDinterwordspacing}{\spaceskip=0pt\relax}
\providecommand{\BIBentryALTinterwordstretchfactor}{4}
\providecommand{\BIBentryALTinterwordspacing}{\spaceskip=\fontdimen2\font plus
\BIBentryALTinterwordstretchfactor\fontdimen3\font minus \fontdimen4\font\relax}
\providecommand{\BIBforeignlanguage}[2]{{%
\expandafter\ifx\csname l@#1\endcsname\relax
\typeout{** WARNING: IEEEtran.bst: No hyphenation pattern has been}%
\typeout{** loaded for the language `#1'. Using the pattern for}%
\typeout{** the default language instead.}%
\else
\language=\csname l@#1\endcsname
\fi
#2}}
\providecommand{\BIBdecl}{\relax}
\BIBdecl

\bibitem{yildirim2024multimodal}
N.~Yildirim, H.~Richardson, M.~T. Wetscherek, J.~Bajwa, J.~Jacob, M.~A. Pinnock, S.~Harris, D.~Coelho De~Castro, S.~Bannur, S.~Hyland \emph{et~al.}, ``Multimodal healthcare ai: identifying and designing clinically relevant vision-language applications for radiology,'' in \emph{Proceedings of the 2024 CHI Conference on Human Factors in Computing Systems}, 2024, pp. 1--22.

\bibitem{yi2025survey}
Z.~Yi, T.~Xiao, and M.~V. Albert, ``A survey on multimodal large language models in radiology for report generation and visual question answering,'' \emph{Information}, vol.~16, no.~2, p. 136, 2025.

\bibitem{mcdonald2015effects}
R.~J. McDonald, K.~M. Schwartz, L.~J. Eckel, F.~E. Diehn, C.~H. Hunt, B.~J. Bartholmai, B.~J. Erickson, and D.~F. Kallmes, ``The effects of changes in utilization and technological advancements of cross-sectional imaging on radiologist workload,'' \emph{Academic radiology}, vol.~22, no.~9, pp. 1191--1198, 2015.

\bibitem{petinaux2011accuracy}
B.~Petinaux, R.~Bhat, K.~Boniface, and J.~Aristizabal, ``Accuracy of radiographic readings in the emergency department,'' \emph{The American journal of emergency medicine}, vol.~29, no.~1, pp. 18--25, 2011.

\bibitem{wang2023r2gengpt}
Z.~Wang, L.~Liu, L.~Wang, and L.~Zhou, ``R2gengpt: Radiology report generation with frozen llms,'' \emph{Meta-Radiology}, vol.~1, no.~3, p. 100033, 2023.

\bibitem{srivastav2024maira}
S.~Srivastav, M.~Ranjit, F.~P{\'e}rez-Garc{\'\i}a, K.~Bouzid, S.~Bannur, D.~C. Castro, A.~Schwaighofer, H.~Sharma, M.~Ilse, V.~Salvatelli \emph{et~al.}, ``Maira at rrg24: A specialised large multimodal model for radiology report generation,'' in \emph{Proceedings of the 23rd Workshop on Biomedical Natural Language Processing}, 2024, pp. 597--602.

\bibitem{xia2024cares}
P.~Xia, Z.~Chen, J.~Tian, Y.~Gong, R.~Hou, Y.~Xu, Z.~Wu, Z.~Fan, Y.~Zhou, K.~Zhu \emph{et~al.}, ``Cares: A comprehensive benchmark of trustworthiness in medical vision language models,'' \emph{Advances in Neural Information Processing Systems}, vol.~37, pp. 140\,334--140\,365, 2024.

\bibitem{abdullah2025multi}
M.~Abdullah, I.~Mansoor, V.~F. Rey, and M.~M. Fraz, ``A multi-llm pipeline for retrieval-grounded, bloom’s taxonomy-aligned question generation,'' in \emph{2025 5th International Conference on Digital Futures and Transformative Technologies (ICoDT2)}.\hskip 1em plus 0.5em minus 0.4em\relax IEEE, 2025, pp. 1--6.

\bibitem{jain2021radgraph}
S.~Jain, A.~Agrawal, A.~Saporta, S.~Q. Truong, D.~N. Duong, T.~Bui, P.~Chambon, Y.~Zhang, M.~P. Lungren, A.~Y. Ng \emph{et~al.}, ``Radgraph: Extracting clinical entities and relations from radiology reports,'' \emph{arXiv preprint arXiv:2106.14463}, 2021.

\bibitem{xia2024rule}
P.~Xia, K.~Zhu, H.~Li, H.~Zhu, Y.~Li, G.~Li, L.~Zhang, and H.~Yao, ``Rule: Reliable multimodal rag for factuality in medical vision language models,'' in \emph{Proceedings of the 2024 Conference on Empirical Methods in Natural Language Processing}, 2024, pp. 1081--1093.

\bibitem{kim2024mdagents}
Y.~Kim, C.~Park, H.~Jeong, Y.~S. Chan, X.~Xu, D.~McDuff, H.~Lee, M.~Ghassemi, C.~Breazeal, and H.~W. Park, ``Mdagents: An adaptive collaboration of llms for medical decision-making,'' \emph{Advances in Neural Information Processing Systems}, vol.~37, pp. 79\,410--79\,452, 2024.

\bibitem{li2024mmedagent}
B.~Li, T.~Yan, Y.~Pan, J.~Luo, R.~Ji, J.~Ding, Z.~Xu, S.~Liu, H.~Dong, Z.~Lin \emph{et~al.}, ``Mmedagent: Learning to use medical tools with multi-modal agent,'' in \emph{Findings of the Association for Computational Linguistics: EMNLP 2024}, 2024, pp. 8745--8760.

\bibitem{mansoor2026reasoning}
I.~Mansoor, M.~Abdullah, M.~D. Rizwan, and M.~M. Fraz, ``Reasoning with large language models in medicine: a systematic review of techniques, challenges and clinical integration,'' \emph{Health Information Science and Systems}, vol.~14, no.~1, p.~6, 2026.

\bibitem{zhang2023biomedclip}
S.~Zhang, Y.~Xu, N.~Usuyama, H.~Xu, J.~Bagga, R.~Tinn, S.~Preston, R.~Rao, M.~Wei, N.~Valluri \emph{et~al.}, ``Biomedclip: a multimodal biomedical foundation model pretrained from fifteen million scientific image-text pairs,'' \emph{arXiv preprint arXiv:2303.00915}, 2023.

\bibitem{yao2022react}
S.~Yao, J.~Zhao, D.~Yu, N.~Du, I.~Shafran, K.~R. Narasimhan, and Y.~Cao, ``React: Synergizing reasoning and acting in language models,'' in \emph{The eleventh international conference on learning representations}, 2022.

\bibitem{sellergren2025medgemma}
A.~Sellergren, S.~Kazemzadeh, T.~Jaroensri, A.~Kiraly, M.~Traverse, T.~Kohlberger, S.~Xu, F.~Jamil, C.~Hughes, C.~Lau \emph{et~al.}, ``Medgemma technical report,'' \emph{arXiv preprint arXiv:2507.05201}, 2025.

\bibitem{demner2016preparing}
D.~Demner-Fushman, M.~D. Kohli, M.~B. Rosenman, S.~E. Shooshan, L.~Rodriguez, S.~Antani, G.~R. Thoma, and C.~J. McDonald, ``Preparing a collection of radiology examinations for distribution and retrieval,'' \emph{Journal of the American Medical Informatics Association}, vol.~23, no.~2, pp. 304--310, 2016.

\bibitem{liu2024factual}
K.~Liu, Z.~Ma, M.~Liu, Z.~Jiao, X.~Kang, Q.~Miao, and K.~Xie, ``Factual serialization enhancement: A key innovation for chest x-ray report generation,'' \emph{arXiv preprint arXiv:2405.09586}, 2024.

\bibitem{banerjee2005meteor}
S.~Banerjee and A.~Lavie, ``Meteor: An automatic metric for mt evaluation with improved correlation with human judgments,'' in \emph{Proceedings of the acl workshop on intrinsic and extrinsic evaluation measures for machine translation and/or summarization}, 2005, pp. 65--72.

\bibitem{zheng2023judging}
L.~Zheng, W.-L. Chiang, Y.~Sheng, S.~Zhuang, Z.~Wu, Y.~Zhuang, Z.~Lin, Z.~Li, D.~Li, E.~Xing \emph{et~al.}, ``Judging llm-as-a-judge with mt-bench and chatbot arena,'' \emph{Advances in neural information processing systems}, vol.~36, pp. 46\,595--46\,623, 2023.

\bibitem{li2023llava}
C.~Li, C.~Wong, S.~Zhang, N.~Usuyama, H.~Liu, J.~Yang, T.~Naumann, H.~Poon, and J.~Gao, ``Llava-med: Training a large language-and-vision assistant for biomedicine in one day,'' \emph{Advances in Neural Information Processing Systems}, vol.~36, pp. 28\,541--28\,564, 2023.

\end{thebibliography}

\end{document}